\documentclass[sigconf,nonacm]{acmart}

\usepackage{amsmath}
\usepackage{booktabs}
\usepackage{cleveref}
\usepackage{graphicx}
\usepackage{microtype}
\usepackage{multirow}
\usepackage{tabularx}
\usepackage{xcolor}
\usepackage{tikz}
\usetikzlibrary{
  arrows.meta,
  backgrounds,
  calc,
  positioning
}
\usepackage{fontawesome5}
\newcommand{\boxTitle}{\sffamily\bfseries\scriptsize}
\newcommand{\boxText}{\sffamily\tiny}

\graphicspath{{figures/}}

\title[FinEvolveKG: Controlled Prompt Evolution]{FinEvolveKG: Controlled and Attributable Prompt
Evolution for Financial Knowledge Graph Construction}

\author{Abhinav Arun}
\authornote{Both authors contributed equally to this research.}
\affiliation{%
  \institution{Domyn}
  \city{New York}
  \country{USA}
}
\email{abhinav.arun@domyn.com}

\author{Jayita Bhattacharyya}
\authornotemark[1]
\affiliation{%
  \institution{Domyn}
  \city{Bengaluru}
  \country{India}
}
\email{jayita.bhattacharyya@domyn.com}

\author{Fabrizio Dimino}
\affiliation{%
  \institution{Domyn}
  \city{New York}
  \country{USA}
}
\email{fabrizio.dimino@domyn.com}

\author{Bhaskarjit Sarmah}
\affiliation{%
  \institution{Domyn}
  \city{Gurugram}
  \country{India}
}
\email{bhaskarjit.sarmah@domyn.com}

\author{Stefano Pasquali}
\affiliation{%
  \institution{Domyn}
  \city{New York}
  \country{USA}
}
\email{stefano.pasquali@domyn.com}

\begin{document}

\begin{abstract}
Financial institutions construct knowledge graphs (KGs) from shared SEC 10-Ks to serve risk analysis, portfolio management, and investment research, whose analytical needs differ. Extraction must adapt to evolving needs without fragmenting
shared grounding, typing, and temporal rules. Hand-authored persona prompts scale poorly, and
persona-specific fine-tuning adds cost. Weight-free prompt optimization offers an alternative, but
unconstrained evolution of a monolithic prompt can rewrite shared rules alongside persona guidance,
obscuring changes and complicating review. To bound this change surface, we introduce
\textbf{FinEvolveKG}, a KG-construction harness that separates a reusable general block from a
persona-policy block and mechanically confines GEPA, a reflective prompt optimizer, to the
designated block. During persona adaptation, the approved general block remains byte-stable and
every optimizer edit is localized to a separately versioned persona policy. A dual objective
balances generic temporal-KG quality with persona-relevant facts, motifs, and multi-hop
connectivity. We evaluate five mutation scopes on ticker-disjoint 10-K splits, with three seeds for
the persona-conditioned arms (v3--v5) across three personas. On the three-seed mean, the complete two-stage
block-constrained workflow (v4) improves over the frozen prompt baseline (v1) by 19.4--30.9\% and
over the tested unconstrained full-prompt workflow (v5) by 2.2--6.4\%, leading v5 on both objective
components and winning all nine paired-seed comparisons while remaining the most seed-stable arm
in every persona. Without further optimization, v4 also exceeds v5 across all evaluated
out-of-distribution (OOD) shifts. An SME audit supports using the generated triplets as ground
truth; a blinded two-SME audit favors v4 over v5 in 70.1\% of contrast-enriched judgments. These results show
that mechanically attributable evolution can coexist with strong empirical performance, without
attributing the gains causally to the boundary.
\end{abstract}


\maketitle

\section{Introduction}
\label{sec:intro}

Financial institutions rarely use an information-extraction system for one homogeneous consumer.
The same SEC 10-K may support a risk function tracing litigation, regulation, and downside
transmission; a portfolio team monitoring performance drivers and forward guidance; and a research
team mapping products, segments, partners, and competitors. All three require source-grounded
facts, consistent entity and relation types, and correct temporal scope, yet they value different
subgraphs and different paths through the evidence. As analytical mandates, downstream
applications, and filing distributions change, the extraction policy must evolve without
fragmenting the evidence rules shared across the institution.

This creates two linked requirements: \emph{evidence integrity} and \emph{analytical utility}.
Existing financial KG pipelines and evaluation frameworks emphasize generic properties such as
faithfulness, precision, relevance, comprehensiveness, schema validity, and temporal
quality~\cite{Arun2025FinReflectKGAC,dimino2025finreflectkgevalbench}. Those properties are necessary,
but a graph can satisfy them while omitting or disconnecting the structures a particular analyst
uses. We therefore treat persona-relevant fact recovery, topicality, and motif connectivity as
optimization-time structural proxies for analyst usability, while retaining generic KG quality as
a separate objective component. These proxies complement rather than replace downstream
question-answering or retrieval evaluation. For financial deployment, both components must remain
visible: improving one analyst's view should not silently weaken the shared evidence contract.

Hand-authoring a prompt for every persona creates costly, independently maintained forks,
while persona-specific fine-tuning adds training, versioning, and deployment overhead. Prompt
evolution offers a weight-free middle path. GEPA reflects over sampled LM-program outputs and
evaluator feedback to propose instruction updates~\cite{agrawal2025gepa}; FinReflectKG provides a
finance-specific precedent for reflection-driven KG construction and
evaluation~\cite{Arun2025FinReflectKGAC}. We extend this principle from refining individual graph
outputs to evolving the reusable instructions that govern extraction. Weight-free evolution,
however, is not inherently controlled. In a monolithic prompt, generic grounding, typing, and
temporal rules can change together with persona priorities. Although the resulting diff records
what changed, it does not assign each edit to a stable functional concern.

This ambiguity matters in finance because extraction policies are reused across filings, analysts,
and downstream applications, and production changes must be reviewable. We introduce
\textbf{FinEvolveKG} to make the change boundary explicit. It separates a reusable General
KG-Construction Block from a Use-Case \& Persona Block and mechanically exposes only the designated
block to GEPA. During persona adaptation, the approved general block remains a named, byte-stable
dependency, so optimizer edits are localized to the persona policy and can be reviewed or released
independently. The mechanical guarantee is deliberately narrow: it establishes
\emph{edit attribution}, not which phrase
caused a score change or whether end-to-end behavior remained invariant. Persona instructions can
still affect graph density, typing, and temporal output. FinEvolveKG therefore combines generic and
persona objectives, structured financial-KG feedback, ticker-aware sampling, and diversity-aware
search, and evaluates five mutation-scope arms (\Cref{tab:arms}) to characterize behavior under different editable
surfaces. We study three questions:
\begin{enumerate}
  \item Does the complete FinEvolveKG workflow (v4) deliver strong generic KG quality
        and persona utility relative to a frozen prompt (v1) and the tested unconstrained workflow
        (v5)?
  \item What edit-control, attribution, and reviewability properties follow from mechanically
        bounded persona adaptation, and what extraction policies emerge?
  \item How well do the optimized prompt artifacts generalize across sector/temporal shifts
        without further optimization?
\end{enumerate}

Across three optimization seeds, v4 improves over both v1 and v5 for every persona and wins all
nine paired-seed comparisons. Without further optimization, canonical seed-42 v4 also exceeds v5
across all evaluated out-of-distribution (OOD) shifts. These
descriptive results support coexistence of control and performance rather than universal dominance. By construction, each v4 artifact pairs a fixed general-block version with an
independently reviewable persona-block diff.

Our key contributions are:
\begin{itemize}
  \item \textbf{Finance-specific formulation.} We formulate persona adaptation as evolving one
        financial KG-construction pipeline toward different financial personas while retaining shared
        evidence-integrity requirements.
  \item \textbf{Controlled architecture.} We introduce a mechanically enforced two-block design
        that confines edits and packages block versions, diffs, and scores as a
        reviewable artifact.
  \item \textbf{Persona-aware optimization.} We develop a GEPA harness that combines generic
        KG quality with persona facts, topicality, and motif connectivity, using structured
        feedback and diversity-aware search over ticker-stratified data.
  \item \textbf{Empirical evaluation.} Across five arms on ticker-disjoint splits and three
        personas (v3--v5 over three seeds; v1 and v2 fixed by design), v4 leads v1 and v5 in-domain
        and, at the canonical seed-42, exceeds v5 in all 15 OOD comparisons. A stratified SME audit
        supports the generated triplet sets as optimization targets, and a blinded two-SME audit
        favors v4 over v1 and v5 on contrast-enriched cases.
\end{itemize}

\section{Related Work}
\label{sec:related}

\paragraph{Financial KG construction and evaluation.}
LLM-based KG systems increasingly separate extraction, schema induction, canonicalization, and
consolidation. EDC formalizes an extract--define--canonicalize pipeline~\cite{zhang2024edc};
AutoSchemaKG induces schemas at scale~\cite{bai2025autoschemakg}; and KGGen emphasizes information
retention and graph consolidation~\cite{mo2025kggen}. Financial systems construct dynamic graphs
from news or event-enhanced graphs from research text~\cite{li2024findkg,li2025finkario}.
FinReflectKG targets source-grounded temporal graphs from SEC filings, with EvalBench providing
multi-dimensional assessment~\cite{Arun2025FinReflectKGAC,dimino2025finreflectkgevalbench}. Its
reflection-driven agentic loop is the closest financial-domain inspiration for our use of GEPA;
FinEvolveKG shifts the reflective search target to a mechanically bounded prompt block and studies
controlled adaptation of one extraction pipeline for multiple consumers. Hyper-KGGen is the
closest KG-specific self-improvement framework: it evolves a global skill library from
stability-based feedback for document-level hypergraph extraction~\cite{huang2026hyperkggen}.
FinEvolveKG instead targets typed temporal financial graphs and separates reusable construction
rules from persona policies with different mutation and review lifecycles.

\paragraph{Prompt and context optimization.}
APE generates and selects instructions, OPRO uses an LLM as the optimizer, and ProTeGi and TextGrad
turn natural-language feedback into textual update signals~\cite{zhou2022ape,yang2023opro,
pryzant2023protegi,yuksekgonul2024textgrad}. DSPy and MIPRO extend optimization from single prompts
to instructions and demonstrations within LM programs~\cite{khattab2023dspy,opsahlong2024mipro}.
GEPA combines reflection over sampled program executions with evolutionary, Pareto-based candidate
selection~\cite{agrawal2025gepa}. ACE and SkillOpt similarly improve external context artifacts
without modifying task-model weights~\cite{zhang2025ace,yang2026skillopt}. FinEvolveKG adopts this
weight-free adaptation paradigm but makes the permitted mutation region a financial system-design
and experimental variable.

\paragraph{Controlled and modular prompt optimization.}
Modular Prompt Optimization applies section-local textual gradients within a fixed prompt schema,
while Adaptive Prompt Structure Factorization discovers semantic factors and uses interventional
factor-level scoring to select one for update~\cite{sharma2026modularprompt,
liu2026adaptivepromptfactorization}. Both address limitations of monolithic rewriting.
FinEvolveKG's distinction is a predefined, domain-owned boundary between shared financial-KG rules
and persona policy, mechanical freezing through a separate program input, and sequential reuse of
an already optimized shared block. The resulting guarantee concerns where edits can occur; effects
on composed-program behavior remain empirical.

\paragraph{Persona-relative KG utility.}
Intrinsic graph scores do not establish usefulness to a downstream consumer. KGrEaT evaluates KGs
through applications~\cite{heist2023kgreat}, while KGCQual measures interpretable construction
properties including entity completeness, resolution, connectivity, and predicate
preservation~\cite{misra2026kgcqual}. In finance, FinReflectKG-MultiHop shows that recurring
two- and three-hop paths support analyst-style questions with compact evidence
contexts~\cite{arun2025finreflectkgmultihop}. FinEvolveKG moves this structural signal inside the
optimization loop: persona motifs and connectivity reward graphs likely to support a consumer's
reasoning pattern, but remain proxies rather than substitutes for downstream task validation.

\section{Problem Formulation}
\label{sec:problem}

Let $d$ be a filing chunk and $\rho$ a consumer persona. The extractor produces a typed temporal
graph
\begin{equation}
G_\rho(d)=f_{\theta}\!\left(d;\mathcal{B}_{g},\mathcal{B}_{\rho}\right),
\end{equation}
where $\mathcal{B}_{g}$ contains general construction instructions and
$\mathcal{B}_{\rho}$ contains use-case and persona instructions. Each triplet has head and tail
entities and types, a relation, start and end timestamps, and a timestamp-source flag.

\paragraph{Two instruction layers.}
$\mathcal{B}_{g}$ defines the canonical evidence behavior: grounding, permitted schema, naming and
deduplication, temporal treatment, and output structure. $\mathcal{B}_{\rho}$ defines the utility
view: business purpose, priority entity and relation types, and characteristic graph motifs. A
persona changes what supported information is emphasized or materialized; it does not authorize a
different financial truth.

\paragraph{Edit guarantees and behavioral effects.}
The block boundary provides two mechanical guarantees: GEPA can modify only the designated block,
and every resulting textual edit is attributable to that block. These guarantees concern the
prompt artifact, not the extracted graph. Because the LLM interprets the general and persona
instructions jointly, persona-block changes may still affect generic extraction behavior. We
therefore measure generic quality after adaptation rather than assume behavioral preservation.

\paragraph{Optimization objective.}
For general quality $\mathcal{Q}_{g}$ and persona utility $\mathcal{Q}_{\rho}$, persona adaptation
maximizes
\begin{equation}
\mathcal{Q}(G_\rho)=
w\,\mathcal{Q}_{g}(G_\rho)+(1-w)\,\mathcal{Q}_{\rho}(G_\rho), \qquad w=0.5.
\label{eq:objective}
\end{equation}
Including $\mathcal{Q}_{g}$ makes generic degradation visible to candidate selection and penalizes
it in the blended score. This is a soft trade-off rather than a hard preservation constraint: a
sufficiently large persona gain may still offset lower generic quality, so behavioral preservation
remains empirical.

\section{FinEvolveKG Methodology}
\label{sec:method}

FinEvolveKG combines a bounded instruction architecture, teacher-generated reference graphs and
dual evaluation, and a reflective search harness. We describe these components in the order needed
to interpret the five mutation scopes.

\subsection{Mechanically Bounded Evolution}
\label{sec:blocks}

GEPA rewrites a DSPy predictor's \texttt{signature.instructions}. FinEvolveKG places the mutable
block in that field and supplies the frozen block through
\texttt{fixed\_guidelines}, an input to every call. For each extraction call, the task model
receives both blocks. However, only the block assigned to \texttt{signature.instructions} is
represented in GEPA's candidate state; \texttt{fixed\_guidelines} is injected unchanged into every
call. Assigning the general or persona block to the mutable field therefore changes the permitted
edit surface while holding the extractor, data path, and output schema fixed.

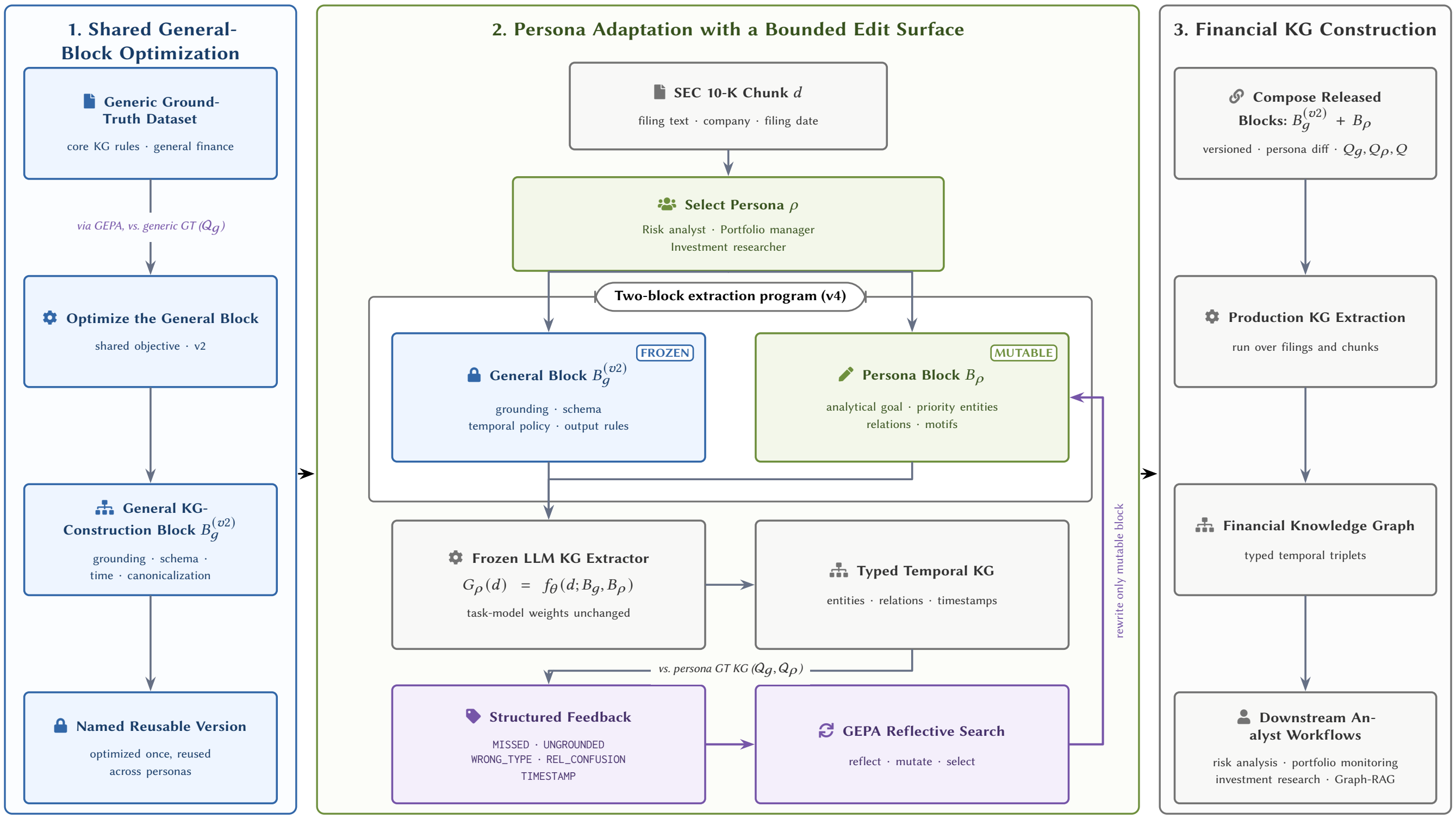
\begin{figure*}[t]
\centering

\resizebox{0.99\textwidth}{!}{%
\begin{tikzpicture}[
  font=\sffamily\scriptsize,
  line cap=round,
  line join=round,
  flow/.style={
    -{Stealth[length=2.1mm,width=1.5mm]},
    draw=arrowgray,
    line width=0.8pt
  },
  stageflow/.style={
    -{Stealth[length=2.4mm,width=1.8mm]},
    draw=arrowgray,
    line width=1.25pt
  },
  feedbackflow/.style={
    -{Stealth[length=2.1mm,width=1.5mm]},
    draw=purpleStroke,
    line width=0.9pt
  },
  basebox/.style={
    draw,
    rounded corners=2pt,
    line width=0.8pt,
    align=center,
    inner xsep=6pt,
    inner ysep=6pt,
    minimum height=1.15cm
  },
  general/.style={
    basebox,
    fill=generalFill,
    draw=generalStroke,
    text=generalText
  },
  persona/.style={
    basebox,
    fill=personaFill,
    draw=personaStroke,
    text=personaText
  },
  execute/.style={
    basebox,
    fill=executeFill,
    draw=executeStroke,
    text=executeText
  },
  optimize/.style={
    basebox,
    fill=purpleFill,
    draw=purpleStroke,
    text=purpleText
  },
  review/.style={
    basebox,
    fill=reviewFill,
    draw=reviewStroke,
    text=reviewText
  },
  stageTitle/.style={
    font=\sffamily\bfseries\small,
    align=center
  }
]


\definecolor{generalFill}{HTML}{EEF5FF}
\definecolor{generalStroke}{HTML}{3267A8}
\definecolor{generalText}{HTML}{173B66}
\definecolor{stageOneFill}{HTML}{FBFDFF}

\definecolor{personaFill}{HTML}{F3F7EA}
\definecolor{personaStroke}{HTML}{6E8F3D}
\definecolor{personaText}{HTML}{36511D}
\definecolor{stageTwoFill}{HTML}{FCFDF9}

\definecolor{purpleFill}{HTML}{F4EFFB}
\definecolor{purpleStroke}{HTML}{7655A8}
\definecolor{purpleText}{HTML}{432E68}

\definecolor{reviewFill}{HTML}{FFF6EA}
\definecolor{reviewStroke}{HTML}{C57922}
\definecolor{reviewText}{HTML}{6D3D0C}
\definecolor{stageThreeFill}{HTML}{FFFDFA}

\definecolor{executeFill}{HTML}{F8F8F7}
\definecolor{executeStroke}{HTML}{737373}
\definecolor{executeText}{HTML}{303030}
\definecolor{stageFourFill}{HTML}{FCFCFC}

\definecolor{arrowgray}{HTML}{667085}

%

\begin{scope}[on background layer]

  \filldraw[
    fill=stageOneFill,
    draw=generalStroke,
    line width=0.75pt,
    rounded corners=3pt
  ]
  (0,0) rectangle (4.20,-11.65);

  \filldraw[
    fill=stageTwoFill,
    draw=personaStroke,
    line width=0.75pt,
    rounded corners=3pt
  ]
  (4.50,0) rectangle (16.35,-11.65);

  \filldraw[
    fill=stageFourFill,
    draw=executeStroke,
    line width=0.75pt,
    rounded corners=3pt
  ]
  (16.65,0) rectangle (20.85,-11.65);

\end{scope}


\node[
  stageTitle,
  text=generalText,
  text width=3.79cm,
  anchor=north
] at (2.10,-0.12)
{1.\ Shared General-Block Optimization};

\node[
  stageTitle,
  text=personaText,
  text width=11.03cm,
  anchor=north
] at (10.43,-0.12)
{2.\ Persona Adaptation with a Bounded Edit Surface};

\node[
  stageTitle,
  text=executeText,
  text width=3.8cm,
  anchor=north
] at (18.75,-0.12)
{3.\ Financial KG Construction};


\node[
  general,
  text width=3.22cm,
  minimum height=1.60cm
] (A) at (2.10,-1.70)
{
  {\boxTitle
   \textcolor{generalStroke}{\faFile}\;
   Generic Ground-Truth Dataset}
  \\[4pt]
  {\boxText core KG rules $\cdot$ general finance}
};

\node[
  general,
  text width=3.22cm,
  minimum height=1.60cm
] (B) at (2.10,-4.70)
{
  {\boxTitle
   \textcolor{generalStroke}{\faCog}\;
   Optimize the General Block}
  \\[4pt]
  {\boxText shared objective $\cdot$ v2}
};

\node[
  general,
  text width=3.22cm,
  minimum height=1.60cm
] (C) at (2.10,-7.70)
{
  {\boxTitle
   \textcolor{generalStroke}{\faSitemap}\;
   General KG-Construction Block $B_g^{(v2)}$}
  \\[4pt]
  {\boxText grounding $\cdot$ schema $\cdot$ time $\cdot$ canonicalization}
};

\node[
  general,
  text width=3.22cm,
  minimum height=1.60cm
] (D) at (2.10,-10.70)
{
  {\boxTitle
   \textcolor{generalStroke}{\faLock}\;
   Named Reusable Version}
  \\[5pt]
  {\boxText
   optimized once, reused\\
   across personas}
};

\draw[flow] (A.south) -- (B.north);
\draw[flow] (B.south) -- (C.north);
\draw[flow] (C.south) -- (D.north);

\node[
  fill=stageOneFill,
  inner xsep=3pt,
  font=\sffamily\itshape\tiny,
  text=purpleStroke
] at (2.10,-3.20)
{via GEPA, vs.\ generic GT ($\mathcal{Q}_g$)};


\node[
  execute,
  text width=4.15cm,
  minimum height=1.25cm
] (E) at (10.43,-1.45)
{
  {\boxTitle
   \textcolor{executeStroke}{\faFile}\;
   SEC 10-K Chunk $d$}
  \\[4pt]
  {\boxText filing text $\cdot$ company $\cdot$ filing date}
};

\node[
  persona,
  text width=5.79cm,
  minimum height=1.35cm
] (P) at (10.43,-3.15)
{
  {\boxTitle
   \textcolor{personaStroke}{\faUsers}\;
   Select Persona $\rho$}
  \\[4pt]
  {\boxText
   Risk analyst $\cdot$ Portfolio manager\\
   Investment researcher}
};

\draw[flow] (E.south) -- (P.north);


\begin{scope}[on background layer]
  \filldraw[
    fill=white,
    draw=executeStroke,
    line width=0.7pt,
    rounded corners=2pt
  ]
  (5.25,-4.20) rectangle (15.67,-7.15);
\end{scope}

\node[
  draw=executeStroke,
  fill=white,
  rounded corners=8pt,
  line width=0.7pt,
  inner xsep=8pt,
  inner ysep=3pt,
  font=\sffamily\bfseries\scriptsize
] at (10.46,-4.20)
{Two-block extraction program (v4)};

\node[
  general,
  text width=4.09cm,
  minimum height=1.85cm
] (G) at (7.84,-5.65)
{
  {\boxTitle
   \textcolor{generalStroke}{\faLock}\;
   General Block $B_g^{(v2)}$}
  \\[6pt]
  {\boxText
   grounding $\cdot$ schema\\
   temporal policy $\cdot$ output rules}
};

\node[
  persona,
  text width=4.09cm,
  minimum height=1.85cm
] (PB) at (13.08,-5.65)
{
  {\boxTitle
   \textcolor{personaStroke}{\faPen}\;
   Persona Block $B_\rho$}
  \\[6pt]
  {\boxText
   analytical goal $\cdot$ priority entities\\
   relations $\cdot$ motifs}
};

\node[
  draw=generalStroke,
  fill=white,
  rounded corners=1.5pt,
  line width=0.6pt,
  font=\sffamily\bfseries\tiny,
  text=generalStroke,
  inner sep=1.5pt,
  anchor=north east,
  xshift=-5pt,
  yshift=-5pt
] at (G.north east)
{FROZEN};

\node[
  draw=personaStroke,
  fill=white,
  rounded corners=1.5pt,
  line width=0.6pt,
  font=\sffamily\bfseries\tiny,
  text=personaStroke,
  inner sep=1.5pt,
  anchor=north east,
  xshift=-5pt,
  yshift=-5pt
] at (PB.north east)
{MUTABLE};

\draw[flow] (P.south) -| (G.north);
\draw[flow] (P.south) -| (PB.north);


\node[
  execute,
  text width=4.09cm,
  minimum height=1.85cm
] (X) at (7.84,-8.35)
{
  {\boxTitle
   \textcolor{executeStroke}{\faCog}\;
   Frozen LLM KG Extractor}
  \\[4pt]
  {\scriptsize $G_\rho(d) = f_\theta(d; B_g, B_\rho)$}
  \\[4pt]
  {\boxText task-model weights unchanged}
};

\node[
  execute,
  text width=4.09cm,
  minimum height=1.85cm
] (KG) at (13.08,-8.35)
{
  {\boxTitle
   \textcolor{executeStroke}{\faSitemap}\;
   Typed Temporal KG}
  \\[5pt]
  {\boxText entities $\cdot$ relations $\cdot$ timestamps}
};

\draw[flow]
  (G.south) -- ++(0,-0.24)
  -| (X.north);

\draw[flow]
  (PB.south) -- ++(0,-0.24)
  -| (X.north);

\draw[flow] (X.east) -- (KG.west);


\node[
  optimize,
  text width=4.09cm,
  minimum height=1.70cm
] (FB) at (7.84,-10.65)
{
  {\boxTitle
   \textcolor{purpleStroke}{\faTag}\;
   Structured Feedback}
  \\[5pt]
  {\boxText
   \texttt{MISSED} $\cdot$ \texttt{UNGROUNDED}\\
   \texttt{WRONG\_TYPE} $\cdot$ \texttt{REL\_CONFUSION}\\
   \texttt{TIMESTAMP}}
};

\node[
  optimize,
  text width=4.09cm,
  minimum height=1.70cm
] (GEPA) at (13.08,-10.65)
{
  {\boxTitle
   \textcolor{purpleStroke}{\faSync}\;
   GEPA Reflective Search}
  \\[5pt]
  {\boxText reflect $\cdot$ mutate $\cdot$ select}
};

\draw[flow]
  (KG.south) -- ++(0,-0.30)
  -| (FB.north);

\node[
  fill=stageTwoFill,
  inner xsep=3pt,
  font=\sffamily\itshape\tiny,
  text=executeText
] at (10.46,-9.575)
{vs.\ persona GT KG ($\mathcal{Q}_{g},\mathcal{Q}_{\rho}$)};

\draw[feedbackflow] (FB.east) -- (GEPA.west);

\draw[feedbackflow]
  (GEPA.east) -- ++(0.48,0)
  |- (PB.east);

\node[
  rotate=90,
  font=\sffamily\tiny,
  text=purpleStroke,
  align=center
] at (16.08,-8.15)
{rewrite only mutable block};


\node[
  execute,
  text width=3.35cm,
  minimum height=1.60cm
] (COMP) at (18.75,-1.70)
{
  {\boxTitle
   \textcolor{executeStroke}{\faLink}\;
   Compose Released Blocks: $B_g^{(v2)} + B_\rho$}
  \\[4pt]
  {\boxText versioned $\cdot$ persona diff $\cdot$ $Q_g, Q_\rho, Q$}
};

\node[
  execute,
  text width=3.35cm,
  minimum height=1.60cm
] (PROD) at (18.75,-4.70)
{
  {\boxTitle
   \textcolor{executeStroke}{\faCog}\;
   Production KG Extraction}
  \\[5pt]
  {\boxText run over filings and chunks}
};

\node[
  execute,
  text width=3.35cm,
  minimum height=1.60cm
] (FKG) at (18.75,-7.70)
{
  {\boxTitle
   \textcolor{executeStroke}{\faSitemap}\;
   Financial Knowledge Graph}
  \\[5pt]
  {\boxText typed temporal triplets}
};

\node[
  execute,
  text width=3.35cm,
  minimum height=1.60cm
] (DW) at (18.75,-10.70)
{
  {\boxTitle
   \textcolor{executeStroke}{\faUser}\;
   Downstream Analyst Workflows}
  \\[5pt]
  {\boxText
   risk analysis $\cdot$ portfolio monitoring\\
   investment research $\cdot$ Graph-RAG}
};

\draw[flow] (COMP.south) -- (PROD.north);
\draw[flow] (PROD.south) -- (FKG.north);
\draw[flow] (FKG.south) -- (DW.north);


\draw[
  ->,
  >=Stealth,
  draw=black,
  line width=0.7pt,
  shorten <=1pt,
  shorten >=1pt
]
(4.20,-6.75) -- (4.50,-6.75);

\draw[
  ->,
  >=Stealth,
  draw=black,
  line width=0.7pt,
  shorten <=1pt,
  shorten >=1pt
]
(16.35,-6.75) -- (16.65,-6.75);

\end{tikzpicture}%
}

\caption{FinEvolveKG optimization and deployment lifecycle. v2 creates a named, reusable general
block; v4 freezes that block while GEPA evolves a persona policy from structured KG feedback.
Released blocks carry their version and persona diff and are composed with their decomposed scores
for production KG construction and downstream analyst workflows.}
\label{fig:architecture}
\Description{A three-stage flow diagram. First, the generic ground-truth dataset for SEC 10-K
training chunks optimizes and versions the reusable general KG-construction block. Second, a filing chunk and one of three
financial personas enter a two-block extraction program containing the frozen general block and a
mutable persona block. A frozen language-model extractor produces a typed temporal graph, which is
compared with the corresponding persona ground-truth KG under both component metrics. Structured
error feedback enters GEPA, which rewrites only the persona block. Third, released blocks are
composed -- carrying the general-block version, persona-block version and diff, and generic and
persona scores -- for production KG extraction, yielding a financial knowledge graph consumed by
risk, portfolio, investment-research, and retrieval workflows.}
\end{figure*}

\paragraph{Sequential reuse.}
As summarized in \cref{fig:architecture}, the workflow has three lifecycle stages comprising four
operational steps. First, optimize $\mathcal{B}_{g}$ on the dedicated generic ground-truth dataset
and select the best validation candidate. Second, complete the first stage by recording that
artifact as a named general-block version. Third, for each persona, instantiate an extractor whose
\texttt{fixed\_guidelines} equals that version and expose only $\mathcal{B}_{\rho}$ to GEPA.
Finally, evaluate the composed program on held-out filings and release the block versions with their
scores, persona diff, and source configuration for production KG construction and downstream
analyst workflows. The result is a versioned prompt artifact rather than a fine-tuned checkpoint.
A new persona repeats only the final two steps; changing the general block requires its dependent
persona blocks to be re-evaluated.

\subsection{Ground-Truth Generation and Validation}
\label{sec:gt}

We generate four ground-truth (GT) KG datasets. The dedicated generic dataset applies core
KG-construction instructions with a general financial-user context and supplies the references used
to optimize v2. The other three datasets correspond to risk analysis, portfolio management, and
investment research. Each persona dataset applies the same generic instructions together with
detailed persona instructions. Every chunk in a persona dataset is additionally tagged as
persona-relevant or non-relevant. Regardless of this tag, both $\mathcal{Q}_{g}$ and
$\mathcal{Q}_{\rho}$ are computed for every per-chunk persona KG, and v3--v5 optimization uses all
available persona records without filtering or rebalancing by it; the tag only defines the
secondary persona-relevant subset used for a canonical-run diagnostic. For a given persona, v3, v4, and v5
use exactly the same chunks and reference KGs. Their differences therefore arise from mutation
scope and, for v4, the inherited v2 artifact rather than different corpus slices. Test and OOD
reference KGs are used only for final scoring and are never exposed to GEPA.

A parameterized generation template varies domain, use case, goal, application, and end-user
context without requiring a separately authored extraction template for every persona. Exhaustive
manual construction and triplet-level validation of financial KGs is prohibitively costly at corpus
scale. We compared Claude Sonnet-4-6, GPT-4o, and DeepSeek-V4-Flash on five of the eight 2024 IT
training tickers using the GEPA metric families, excluding self-match F1. Claude led each persona on the equal-weight blend: .939 for risk analysis, .921 for portfolio management,
and .858 for investment research. Its mean blended score across the three personas was .906,
exceeding GPT-4o's .791 and DeepSeek's .690. This relative result motivated generator selection but
does not independently establish reference validity.

We therefore conducted a stratified audit of 50 chunks balanced across five IT tickers and five
major 10-K sections, with two chunks per ticker--section cell, while avoiding administrative
boilerplate. A financial-domain SME blindly evaluated each Claude-generated triplet set against its
source chunk under the generic and three persona configurations, covering 200 chunk-level
evaluations. Each triplet set was rated \emph{acceptable}, \emph{requiring minor
correction}, or \emph{unreliable}. Acceptable required all reported triplets to be grounded and
semantically correct; isolated duplicates or relation/entity-type semantic errors received a minor-correction
rating, whereas multiple grounding, directionality, or fabrication errors were unreliable.
Missing unrelated facts alone did not trigger a downgrade. Verdict counts (acceptable/minor/unreliable) were
43/7/0 for generic, 45/5/0 for risk analysis, 45/3/2 for portfolio management, and 45/4/1 for
investment research---178/19/3 overall. Thus, 98.5\% were acceptable or required only localized
correction; the three unreliable cases involved ungrounded or fabricated content or inverted
critical relations. This stratified audit supports using Claude-generated triplet sets as
optimization targets but does not constitute exhaustive corpus-wide validation.

\subsection{Five Mutation Scopes}
\label{sec:arms}

\Cref{tab:arms} defines five arms evaluated across three personas. v2 optimizes the general block
once on generic GT. v3--v5 use every chunk from the corresponding persona GT and the blended
objective in \cref{eq:objective}; they differ in which blocks GEPA may edit and which general-block
version is fixed.

\begin{table}[t]
\caption{Mutation scopes and optimization conditions.}
\label{tab:arms}
\footnotesize
\setlength{\tabcolsep}{3pt}
\begin{tabular}{p{.55cm}p{1.12cm}p{1.38cm}p{1.28cm}p{1.42cm}}
\toprule
\textbf{Arm} & \textbf{Mutable} & \textbf{Fixed} & \textbf{GT data} &
\textbf{Selection objective} \\
\midrule
v1 & -- & both at seed & -- & no search \\
v2 & general & persona seed & generic & $\mathcal{Q}_{g}$ \\
v3 & persona & general seed & persona (all) & $\mathcal{Q}$ \\
v4 & persona & general v2 & persona (all) & $\mathcal{Q}$ \\
v5 & full prompt & -- & persona (all) & $\mathcal{Q}$ \\
\bottomrule
\end{tabular}
\end{table}

The primary system comparison is v1 versus v4 versus v5. v2 exposes v4's shared-block optimization
stage, whereas v3 provides a persona-only control initialized from the frozen general seed. For each
persona, v3--v5 use matched persona-stage data, evaluation, and budget; v3 and v4 edit only the
persona block, whereas v5 edits the full seed prompt. v4 additionally inherits the v2 general block
and its preceding search, whose cost is amortized across personas. Thus, v4 versus v5 compares
complete workflows rather than causally isolating the block boundary, initialization, or
optimization history.

\subsection{Persona Policies}
\label{sec:personas}

Each persona policy specifies an analytical goal, priority entity and relation types, and a catalog
of one- and two-hop motifs. \Cref{tab:personas} summarizes each persona's functional ownership and
a representative motif. Motifs do not require the extractor to invent missing edges; they reward
the completion of supported paths when both facts occur in the source. The catalogs contain 10
two-hop risk motifs, eight portfolio motifs, and eight research motifs, in addition to one-hop
patterns.

\begin{table}[t]
\caption{Persona ownership: what each persona prioritizes over the same underlying schema.}
\label{tab:personas}
\footnotesize\setlength{\tabcolsep}{3pt}
\begin{tabular}{p{1.7cm}p{1.3cm}p{1.9cm}p{2.3cm}}
\toprule
\textbf{Persona} & \textbf{Owns} & \textbf{Key types / relations} & \textbf{Example motif} \\
\midrule
risk\_analyst & downside \& compliance & \texttt{RISK\_FACTOR}, \texttt{LITIGATION},
\texttt{ORG\_REG}; \texttt{Negatively\_Impacts}, \texttt{Faces}, \texttt{Regulates} &
\texttt{RISK\_FACTOR}$\to$\allowbreak\texttt{Negatively\_Impacts}$\to$\allowbreak
\texttt{FIN\_METRIC} \\
portfolio\_mgr & quantitative performance & \texttt{FIN\_METRIC}, \texttt{FIN\_INST},
\texttt{SEGMENT}; \texttt{Discloses}, \texttt{Guides\_On} &
\texttt{SEGMENT}$\to$\allowbreak\texttt{Discloses}$\to$\allowbreak\texttt{FIN\_METRIC} \\
investment\_\allowbreak researcher & strategic / competitive & \texttt{PRODUCT}, \texttt{COMP},
\texttt{SECTOR}; \texttt{Produces}, \texttt{Partners\_With} &
\texttt{ORG}$\to$\allowbreak\texttt{Partners\_With}$\to$\allowbreak\texttt{COMP} \\
\bottomrule
\end{tabular}
\end{table}

\subsection{Generic Quality and Analyst Salience}
\label{sec:metrics}

\paragraph{Generic quality.}
$\mathcal{Q}_{g}$ is a normalized weighted average of overlap F1 (.35), temporal accuracy (.25),
entity-type accuracy (.15), source support (.15), density (.10), reconstruction (.10), naming
quality (.08), relation specificity (.04), and deduplication (.03). $\mathcal{Q}_{g}$ is computed
for every chunk. Weights are normalized by their active sum (1.25 when complete). If a submetric
cannot be computed for a record, that term is omitted and the remaining
weights are renormalized; the record itself remains in the evaluation. Per chunk, schema/structure
and grounding scores below .90 multiplicatively reduce the result. They are gates rather than near-saturated reward terms, so
format or grounding failures cannot be cheaply offset by semantic gains. Naming consistency is
reported only as a diagnostic. Feedback identifies missed facts, unsupported predictions,
relation/type confusions, timestamp errors, and the largest weighted sub-score gaps.

\paragraph{Analyst salience.}
Let $Y_{\rho}$ denote the persona-relevant triplet and motif view within a per-chunk reference KG.
This graph-level view is distinct from the binary chunk-level relevance tag. The Analyst Salience
Index is
\begin{equation}
\mathcal{Q}_{\rho}
=0.50\,F1_{\rho}+0.30\,C_{\rho}+0.20\,T_{\rho}.
\label{eq:persona}
\end{equation}
$F1_{\rho}$ uses partial-credit semantic matching over the triplets in $Y_{\rho}$.
$C_{\rho}$ rewards recovery of one-hop motifs and realized two-hop paths whose adjacent edges share
the same intermediate entity. For example, a risk path may connect
\textsc{ORG}$\xrightarrow{\textsc{Faces}}$\textsc{RISK\_FACTOR}
$\xrightarrow{\textsc{Negatively\_Impacts}}$\textsc{FIN\_METRIC}; a portfolio path connects a
disclosed segment to its metric; and a research path connects a partner to a product.
$T_{\rho}$ measures density-aware semantic topicality relative to the persona description.
Equation~\eqref{eq:persona} is evaluated for every chunk, irrespective of its relevance tag. If a
component is not applicable to a record---for example, because no reference motif exists---that
term is omitted and the remaining weights are renormalized. The result is then gated when
output schema/structure falls below .70,
grounding falls below .90, or fewer than .90 of predicted chains are backed by matched gold facts.
The last gate prevents an optimizer from earning connectivity credit by joining unsupported edges.
The reported $\mathcal{Q}_{\rho}$ is the mean of these gated per-chunk scores, so it need not equal
the weighted sum of corpus-level mean sub-metrics. Persona recall, persona temporal accuracy, and
persona type accuracy are feedback-only diagnostics; they do not enter the composite or its gates.
This keeps temporal and type correctness in generic quality while still providing targeted
reflection feedback.

\subsection{Reflective Search and Feedback}
\label{sec:search}

FinEvolveKG configures the GEPA loop in four ways. First, mutation is
mechanically routed to a named block as described in \cref{sec:blocks}. Second, each evaluation
returns both a scalar score and structured natural-language feedback. Stable error tags---including
\texttt{MISSED}, \texttt{UNGROUNDED}, \texttt{WRONG\_TYPE}, and \texttt{TIMESTAMP}---precede the largest weighted sub-score gaps and concrete missed or near-match
triplets. Thus, reflection receives a domain-specific diagnosis rather than a single reward.
Third, a ticker-stratified sampler round-robins companies so a minibatch is not dominated by one
filing. Fourth, search includes lineage-aware parent selection, bounded candidate merges, and
no-improvement stopping to limit redundant late-stage rewrites. GEPA normally samples parents by Pareto coverage over validation examples; the reported run adds
lineage diversity to discourage closely related selections. Exact merge and stopping settings are
reported in \cref{tab:setup}. Because these search extensions are not ablated against default GEPA,
we treat them as shared infrastructure rather than attributing arm-level gains to them. For each persona, v3--v5 receive the same three fixed, training-only demonstrations, sampled from different tickers to exhibit persona motifs and connected
two-hop paths. The demonstrations remain outside GEPA's candidate state, so demonstration choice is
matched across arms; GEPA still optimizes and evaluates over all available records.

\section{Experimental Setup}
\label{sec:setup}

\paragraph{Data and splits.}
We define the company universe primarily from S\&P 100 constituents and GICS sector
labels~\cite{spglobal2026sp100}, with KMB additionally included in the Consumer Staples OOD set. Reference graphs
were generated from 2024 SEC 10-K filings for the 20 Information Technology (IT) constituents. The primary partition is
ticker-disjoint: eight tickers for training, four for GEPA validation, and eight for reporting
(Table~\ref{tab:setup}). All chunks from a ticker remain in one split, preventing filing-specific
leakage. The
completed test sets contain
757 portfolio, 761 risk, and 756 research chunks. Of these, 475, 286, and 257, respectively, are
tagged as persona-relevant. These labels define the secondary held-out subsets; they do not remove
any chunks from v3--v5 optimization.

We additionally evaluate the saved programs without further optimization on five OOD slices
(\cref{tab:setup}). To provide heterogeneous rather than representative sector coverage, the sector
shift uses four constituents from each of three unseen 2024 GICS sectors: Financials, Health Care,
and Consumer Staples. The temporal slice uses four 2025 IT filings whose 2024 counterparts span the training,
validation, and held-out splits, isolating a new filing year under mixed company exposure. The
combined slice uses 2025 Financials filings, where both sector and year are unseen during
optimization.

\paragraph{Models and optimization.}
Claude Sonnet-4-6 supplies reference graphs; \texttt{google/gemma-4-31B-it} is the frozen
extractor; and \texttt{deepseek-ai/DeepSeek-V4-Flash} proposes GEPA reflections. \Cref{tab:setup}
summarizes the search configuration. Validation scores plateau within the 3,000-call budget for
most persona--arm runs, consistent with the ticker split providing sufficient signal.

\begin{table}[t]
\caption{Primary experimental configuration.}
\label{tab:setup}
\footnotesize
\setlength{\tabcolsep}{3pt}
\begin{tabularx}{\columnwidth}{@{}>{\raggedright\arraybackslash}p{2.15cm}>{\raggedright\arraybackslash}X@{}}
\toprule
\textbf{Component} & \textbf{Setting} \\
\midrule
Train (IT, 2024) & AAPL, ACN, ADBE, CRM, AVGO, INTC, ORCL, CSCO \\
Validation (IT, 2024) & AMAT, IBM, INTU, LRCX \\
Test (IT, 2024) & AMD, MSFT, NVDA, MU, NOW, PLTR, QCOM, TXN \\
\addlinespace[1pt]
OOD sector (2024) & Financials: AXP, MA, SCHW, V;\newline
Health Care: ABT, AMGN, BMY, ISRG;\newline
Consumer Staples: CL, KMB, KO, MDLZ \\
OOD temporal & AAPL, IBM, MSFT, NVDA (IT, 2025) \\
OOD combined & AXP, MA, SCHW, V (Financials, 2025) \\
\addlinespace[1pt]
Task / reflection LM & Gemma-4-31B-it / DeepSeek-V4-Flash \\
Metric calls & 3,000 per optimized arm \\
Blend weight & $w=0.5$ for v3--v5 \\
Reflection batch & 8, ticker-stratified \\
Candidate selection & lineage diversity; up to 5 merges \\
Early stopping & 20 no-improvement rounds; skip perfect examples \\
Teacher demos & 3 fixed, train-only, persona/chain rich (v3--v5) \\
Matching / temporal & semantic matching; filing-default timestamps only \\
\bottomrule
\end{tabularx}
\end{table}

\paragraph{Reporting.}
Both $\mathcal{Q}_{g}$ and $\mathcal{Q}_{\rho}$ are computed for every test chunk before
aggregation. We report generic, persona, and equally weighted blended scores on the complete test
distribution, measuring end-to-end behavior under natural prevalence. The relevance tag defines
only a secondary canonical-run diagnostic and never filters or rebalances optimization data: all
v3--v5 arms use the complete chunk distribution. Semantic matching is enabled; temporal scoring
uses filing-default timestamps and skips extracted timestamps. In-domain v3--v5 results are
reported over three optimization seeds; v1 is unoptimized and v2 is the single selected shared
artifact reused across personas and seeds. Detailed submetric and OOD analyses use the canonical
seed-42 artifacts. We plan to release supporting experimental artifacts, learning curves, audit materials, and complete seed-level results in a future revision.

\paragraph{Blinded SME output audit.}
To avoid uninformative comparisons between near-identical outputs, we selected 50 held-out
chunks---10 per ticker from five randomly selected IT tickers---from each ticker's top 30\% by
absolute v4--v5 blended-score gap. Selection ignored gap direction: among the resulting 150 chunk--persona cases, 
v4 scored higher in 56.7\% and v5 in 43.3\%, so both directions were well represented. For each chunk and persona, v1, v4, and v5 triplet sets were generated, and
two independent financial-domain SMEs blindly evaluated the v4--v1 and v4--v5 pairs with model scores and arm identities concealed. On this contrast-enriched sample, v4 was
preferred over v1 in 93.0\% of judgments and over v5 in 70.1\%, with substantial inter-rater
agreement (93.3\% raw agreement; Cohen's $\kappa=.78$). Thus, the audit provides independent human
support for v4's output quality when outputs differ meaningfully rather than estimating preference
across all test chunks.

\section{Results}
\label{sec:results}

\subsection{In-Domain Held-Out Ticker Performance}

\Cref{tab:results} reports the completed three-persona test. The primary system-level comparison is
v1 versus v4 versus v5: a frozen prompt, the complete sequential FinEvolveKG workflow, and the
tested unconstrained full-prompt workflow. We retain v2 and v3 to expose the shared-only and
persona-on-seed stages used to construct and interpret v4. The complete test sets contain 761 risk,
757 portfolio, and 756 research chunks. For the complete-workflow comparison, v4 improves over v1 by 27.4\% for risk, 30.9\% for portfolio,
and 19.4\% for research on the mean blended score. Relative to v5, the gains are 6.4\%, 4.3\%, and
2.2\%, respectively. Every optimized arm exceeds v1 on all three reported headline scores in every
full-distribution persona condition. As a secondary diagnostic, the canonical seed-42 v4-over-v5
ordering also holds when the same per-chunk metrics are aggregated over the tagged
persona-relevant subset.

\begin{table}[t]
\caption{Eight-ticker held-out scores over all chunks. v1 is the frozen baseline and v2 the single
selected shared artifact; v3--v5 report mean $\pm$ sample SD over seeds 7, 42, and 123. Bold arm labels
identify the primary comparison; bold values mark the best blended score.}
\label{tab:results}
\footnotesize
\setlength{\tabcolsep}{3pt}
\begin{tabular}{@{}llccc@{}}
\toprule
\textbf{Persona} & \textbf{Arm} & $\mathcal{Q}_{g}$ & $\mathcal{Q}_{\rho}$ & $\mathcal{Q}$ \\
\midrule
\multirow{5}{*}{Risk analyst}
 & \textbf{v1} & .7187 & .1186 & .4186 \\
 & v2 & .8108 & .1678 & .4893 \\
 & v3 & $.8190 \pm .0132$ & $.2078 \pm .0137$ & $.5134 \pm .0135$ \\
 & \textbf{v4} & $.8120 \pm .0000$ & $.2543 \pm .0000$ & $\mathbf{.5331 \pm .0001}$ \\
 & \textbf{v5} & $.8069 \pm .0120$ & $.1955 \pm .0105$ & $.5012 \pm .0034$ \\
\midrule
\multirow{5}{*}{Portfolio manager}
 & \textbf{v1} & .7258 & .2061 & .4659 \\
 & v2 & .8109 & .3570 & .5839 \\
 & v3 & $.8293 \pm .0016$ & $.3848 \pm .0370$ & $.6071 \pm .0179$ \\
 & \textbf{v4} & $.8237 \pm .0000$ & $.3962 \pm .0002$ & $\mathbf{.6099 \pm .0001}$ \\
 & \textbf{v5} & $.8154 \pm .0183$ & $.3541 \pm .0251$ & $.5848 \pm .0212$ \\
\midrule
\multirow{5}{*}{Investment researcher}
 & \textbf{v1} & .7559 & .2945 & .5252 \\
 & v2 & .8344 & .3606 & .5975 \\
 & v3 & $.8195 \pm .0055$ & $.3775 \pm .0029$ & $.5985 \pm .0042$ \\
 & \textbf{v4} & $.8343 \pm .0017$ & $.4203 \pm .0029$ & $\mathbf{.6273 \pm .0023}$ \\
 & \textbf{v5} & $.8219 \pm .0106$ & $.4054 \pm .0195$ & $.6136 \pm .0151$ \\
\bottomrule
\end{tabular}
\end{table}

The equal-budget v3--v5 contrast controls for search effort: both start from the same seed prompt
and share the data, objective, demonstrations, and 3,000-call budget, but v3 freezes the general
block and edits only the persona block, whereas v5 may rewrite the full prompt. On the three-seed
mean, v3 exceeds v5 for risk analysis (.5134 versus .5012) and portfolio management (.6071 versus
.5848), while v5 leads investment research (.6136 versus .5985). Thus, enlarging the editable
surface does not reliably improve the tested finite-budget search, although this comparison is not
a causal estimate of the constraint's effect.

\subsection{Multi-Seed Robustness Check}
\label{sec:multiseed}

Across seeds 7, 42, and 123, with v1 and the selected v2 artifact fixed by design, v4 exceeds v5 in
all nine paired comparisons; the mean blended-score gaps are .0319 for risk analysis, .0252 for
portfolio management, and .0137 for investment research. v4 exceeds v3 in eight of nine
comparisons---the exception is portfolio seed 123---and has the lowest observed blended-score SD in
every persona (.0001, .0001, and .0023, versus v5's .0034, .0212, and .0151). These $n=3$
comparisons are descriptive rather than formal significance tests and measure persona-stage
variation conditional on the selected shared artifact.

\subsection{Objective Components and Submetric Analysis}

While \cref{tab:results} reports multi-seed headline estimates, we use the canonical seed-42
artifacts for diagnostic analysis. For risk analysis, portfolio management, and investment
research, respectively, the v4/v5 generic scores are .8120/.8075, .8237/.8186, and .8333/.8280;
the corresponding persona scores are .2543/.1883, .3960/.3457, and .4186/.4165. Thus, v4 leads both
objective components in all three comparisons, so its blended advantage does not trade away the
shared objective, although the investment-research persona comparison is close. Relative submetric
changes show different patterns: risk has 40.7\% higher topicality but 12.5\% lower F1 and 5.3\%
lower connectivity; portfolio improves all three by 2.9\%, 6.1\%, and 12.5\%; and research improves
F1 and connectivity by 1.6\% and 8.5\%, while topicality is effectively tied (+0.1\%). Across
personas, v4 improves generic temporal accuracy by 2.2--5.9\% and deduplication by 9.6--16.0\%, but
trails specificity by 3.2--8.2\%. Because gates operate per chunk, these corpus-level changes
diagnose trade-offs rather than additively decompose $\mathcal{Q}$.

\subsection{Out-of-Distribution Generalization}
\label{sec:ood}

We apply the canonical seed-42 artifacts without further optimization to the five OOD conditions in
\cref{tab:ood}, each comprising four filings and yielding 415--1,162 evaluated chunks, depending on
the condition and persona. v4 exceeds
v5 in all 15 OOD persona--condition comparisons and ranks first among all five arms in 11. It leads
v5 on both objective components for risk analysis and portfolio management throughout. For
investment research, v4 leads generic quality in all six displayed conditions but persona utility
in only two, so its generic advantage drives the higher blended score. The four exceptions to v4's
best-arm ranking occur in portfolio sector shifts, where the fixed shared-only v2 artifact leads;
v4 remains best under the pure temporal shift. These descriptive results support consistent
v4-over-v5 transfer across the tested sectors and years, not universal best-arm performance or
generalization.

\begin{table*}[!t]
\caption{Blended score $\mathcal{Q}$ for the canonical seed-42 artifacts across in-domain and OOD
evaluations, without further optimization. v1 is unoptimized and v2 is the fixed shared artifact;
seed 42 identifies the v3--v5 optimization runs. Bold marks the best arm.}
\label{tab:ood}
\footnotesize
\setlength{\tabcolsep}{5pt}
\renewcommand{\arraystretch}{0.92}
\begin{tabular}{@{}llcccccc@{}}
\toprule
 & & \multicolumn{1}{c}{\textbf{In-domain}} & \multicolumn{3}{c}{\textbf{Sector shift ('24)}}
 & \multicolumn{1}{c}{\textbf{Temporal shift}} & \multicolumn{1}{c}{\textbf{Sector + temporal}} \\
\cmidrule(lr){3-3}\cmidrule(lr){4-6}\cmidrule(lr){7-7}\cmidrule(lr){8-8}
\textbf{Persona} & \textbf{Arm} & \textbf{IT '24} & \textbf{Financials} & \textbf{Health Care}
 & \textbf{Consumer Staples} & \textbf{IT '25} & \textbf{Financials '25} \\
\midrule
\multirow{5}{*}{Risk analyst}
 & v1 & .4186 & .3458 & .3629 & .3446 & .4021 & .3900 \\
 & v2 & .4893 & .4177 & .4349 & .4261 & .4883 & .4724 \\
 & v3 & .4978 & .3925 & .3683 & .3891 & .4698 & .4383 \\
 & v4 & \textbf{.5331} & \textbf{.4344} & \textbf{.4552} & \textbf{.4427} & \textbf{.5256} & \textbf{.5039} \\
 & v5 & .4979 & .3914 & .4098 & .3980 & .4695 & .4479 \\
\midrule
\multirow{5}{*}{Portfolio manager}
 & v1 & .4659 & .4587 & .4555 & .4709 & .4448 & .4453 \\
 & v2 & .5839 & \textbf{.5743} & \textbf{.5559} & \textbf{.5851} & .5608 & \textbf{.5849} \\
 & v3 & .6047 & .5704 & .5523 & .5642 & .5573 & .5726 \\
 & v4 & \textbf{.6098} & .5488 & .5421 & .5780 & \textbf{.5776} & .5831 \\
 & v5 & .5822 & .5260 & .5164 & .5389 & .5269 & .5309 \\
\midrule
\multirow{5}{*}{Investment researcher}
 & v1 & .5252 & .3944 & .4569 & .4141 & .4806 & .4387 \\
 & v2 & .5975 & .4716 & .5331 & .5059 & .5693 & .5351 \\
 & v3 & .5958 & .4672 & .5209 & .4854 & .5611 & .5267 \\
 & v4 & \textbf{.6260} & \textbf{.4952} & \textbf{.5573} & \textbf{.5362} & \textbf{.5973} & \textbf{.5633} \\
 & v5 & .6223 & .4826 & .5491 & .5155 & .5899 & .5415 \\
\bottomrule
\end{tabular}
\end{table*}

\subsection{Qualitative Prompt Analysis}
\label{sec:qualitative}

Saved prompt artifacts distinguish optimizer-written instructions from attached demonstrations and
show where each arm changes; the excerpts establish edit attribution, not phrase-level causation.

The v2 general block turns v1's broad request for \emph{``ALL relevant temporal triplets that
capture domain-specific knowledge''} into operational rules for line, table-row, and footnote
coverage, entity typing, and grounding. It requires that \emph{``Every entity in a triplet MUST
co-occur with the other entity in the same sentence or directly adjacent sentences''} and organizes
failures as \emph{Low Density}, \emph{Duplicates}, \emph{No Co-occurrence}, \emph{Wrong Entity
Types}, and \emph{Temporal Errors}, mirroring evaluator feedback. v4 reuses this named, byte-stable
general block across all personas, making the location of subsequent persona-specific edits
explicit.

In v4, bounded persona evolution produces distinct policies without rewriting the shared block.
Investment research allows \texttt{Partners\_With} \emph{``ONLY if the text explicitly uses
`partner' or clearly indicates a strategic collaboration.''} Portfolio management identifies a
\emph{``Chain Broken''} failure and directs the extractor to \emph{``extract both hops''} from a
company to a financial metric and from the metric to its reported value. This rule converts
quantitative table rows into connected KG triplets, addressing a table-to-KG extraction challenge
documented by FinReflectKG~\cite{Arun2025FinReflectKGAC}. The risk block remains at its seed value,
showing that the boundary records both substantive persona edits and an explicit no-change outcome.

The v5 artifacts show how unconstrained full-prompt optimization weakens review control by removing
edit separation. All three change the shared temporal rule for point-in-time facts: v2 spans the
default start through the observed month, whereas v5 uses that month as both endpoints. Risk v5 also
replaces v2's \texttt{Discloses} relation for broad risk factors with \texttt{Faces};
investment-research v5 types named equity plans as \texttt{PRODUCT} and connects them with
\texttt{Produces}. These revisions may be useful individually, but their interleaving with persona
edits in a monolithic prompt prevents a reviewer from retaining previously approved temporal,
typing, and relation policies while accepting only the persona adaptation. Only portfolio v5
labels a separate ``Portfolio Manager Lens''; ownership remains implicit in the risk and
investment-research artifacts.

\section{Conclusion}
\label{sec:conclusion}

FinEvolveKG frames persona adaptation of financial KG construction as controlled evolution of a
reusable extraction program. Its two-block design versions a shared general policy and confines
each persona run to a separately reviewable local policy, providing edit control and attribution
without updating model weights. On held-out IT filings, v4 improves over v1 by 27.4\%, 30.9\%, and
19.4\% and over the tested unconstrained v5 workflow by 6.4\%, 4.3\%, and 2.2\% on the three-seed
mean, winning all nine paired comparisons. Applied unchanged, the canonical seed-42 v4 artifacts
also exceed v5 in all 15 OOD comparisons. Artifact analysis shows that v4 preserves a named shared
block, whereas v5 interleaves shared and persona edits. These results show that reviewable edit
control can coexist with strong empirical performance under the tested conditions; they do not
establish that the boundary causes the gains or eliminate the need for output validation. By
producing versioned prompt artifacts rather than persona-specific checkpoints or full-prompt forks,
FinEvolveKG offers financial institutions a lightweight way to evolve a shared KG-construction
harness as analyst needs and filing distributions evolve while preserving reviewable ownership of
production changes.

\section{Limitations and Future Work}
\label{sec:limitations}

The multi-seed analysis varies persona-stage optimization over $n=3$ seeds while holding the
selected v2 shared artifact fixed. This matches the intended reuse of one shared block but leaves
variability in shared-block selection unmeasured. OOD evaluation uses only the canonical seed-42
artifacts and a limited set of sectors and filing years; future work will expand both sector and
temporal coverage. Reference-KG validity was assessed by one SME, while the two-SME
output-preference audit targeted contrast-enriched cases; larger stratified audits will broaden both
validation layers.

Analyst usability is currently proxied through persona-relevant schema elements and KG motifs
rather than downstream decisions. Future work will evaluate downstream task utility directly using
Graph-RAG agents on representative analyst-oriented QA benchmarks and test sensitivity to the
fixed blend weight. Finally, the
current boundary guarantees edit location, not semantic separation or causal performance effects;
future work will investigate semantic ownership constraints that prevent mutable blocks from
restating or overriding shared guidance.

\bibliographystyle{ACM-Reference-Format}
\bibliography{references}

@misc{spglobal2026sp100,
  author = {{S\&P Dow Jones Indices}},
  title  = {{S\&P 100}},
  year   = {2026},
  url    = {https://www.spglobal.com/spdji/en/indices/equity/sp-100/},
  note   = {Accessed August 1, 2026},
}

@article{Arun2025FinReflectKGAC,
  title   = {{FinReflectKG}: Agentic Construction and Evaluation of Financial Knowledge Graphs},
  author  = {Arun, Abhinav and Dimino, Fabrizio and Agarwal, Tejas Prakash and Sarmah, Bhaskarjit and Pasquali, Stefano},
  journal = {Proceedings of the 6th ACM International Conference on AI in Finance},
  year    = {2025},
  doi     = {10.1145/3768292.3770363},
}

@article{arun2025finreflectkgmultihop,
  author  = {Arun, Abhinav and Harsh, Reetu Raj and Sarmah, Bhaskarjit and Pasquali, Stefano},
  title   = {{FinReflectKG--MultiHop}: Financial QA Benchmark for Reasoning with Knowledge Graph Evidence},
  journal = {arXiv preprint},
  year    = {2025},
  url     = {https://arxiv.org/abs/2510.02906},
}

@article{dimino2025finreflectkgevalbench,
  author  = {Dimino, Fabrizio and Arun, Abhinav and Sarmah, Bhaskarjit and Pasquali, Stefano},
  title   = {{FinReflectKG--EvalBench}: Benchmarking Financial KG with Multi-Dimensional Evaluation},
  journal = {arXiv preprint},
  year    = {2025},
  url     = {https://arxiv.org/abs/2510.05710},
}

@inproceedings{agrawal2025gepa,
  author    = {Agrawal, Lakshya A and Tan, Shangyin and Soylu, Dilara and Ziems, Noah and
               Khare, Rishi and Opsahl-Ong, Krista and Singhvi, Arnav and Shandilya, Herumb and
               Ryan, Michael J and Jiang, Meng and Potts, Christopher and Sen, Koushik and
               Dimakis, Alexandros G. and Stoica, Ion and Klein, Dan and Zaharia, Matei and
               Khattab, Omar},
  title     = {{GEPA}: Reflective Prompt Evolution Can Outperform Reinforcement Learning},
  booktitle = {Proceedings of the International Conference on Learning Representations (ICLR)},
  year      = {2026},
  url       = {https://openreview.net/forum?id=RQm2KQTM5r},
}

@article{sharma2026modularprompt,
  title   = {Modular Prompt Optimization: Optimizing Structured Prompts with Section-Local Textual Gradients},
  author  = {Sharma, Prith and Henley, Austin Z.},
  journal = {arXiv preprint},
  year    = {2026},
  url     = {https://arxiv.org/abs/2601.04055},
}

@inproceedings{liu2026adaptivepromptfactorization,
  title     = {Adaptive Prompt Structure Factorization: A Framework for Self-Discovering and Optimizing Compositional Prompt Programs},
  author    = {Liu, Haoyue and Wang, Zhichao and Guo, Yongxin and Shou, Haoran and Tang, Xiaoying},
  booktitle = {Proceedings of the 64th Annual Meeting of the Association for Computational Linguistics (Volume 1: Long Papers)},
  pages     = {11690--11714},
  publisher = {Association for Computational Linguistics},
  year      = {2026},
  doi       = {10.18653/v1/2026.acl-long.537},
}

@inproceedings{huang2026hyperkggen,
  title     = {{Hyper-KGGen}: A Skill-Driven Knowledge Extractor for High-Quality Knowledge Hypergraph Generation},
  author    = {Huang, Rizhuo and Feng, Yifan and Xue, Rundong and Ying, Shihui and Yong, Jun-Hai and Shi, Chuan and Du, Shaoyi and Gao, Yue},
  booktitle = {Proceedings of the 32nd ACM SIGKDD Conference on Knowledge Discovery and Data Mining V.2},
  year      = {2026},
  doi       = {10.1145/3770855.3818198},
}

@article{zhang2025ace,
  title   = {Agentic Context Engineering: Evolving Contexts for Self-Improving Language Models},
  author  = {Zhang, Qizheng and Hu, Changran and Upasani, Shubhangi and Ma, Boyuan and Hong, Fenglu and Kamanuru, Vamsidhar and Rainton, Jay and Wu, Chen and Ji, Mengmeng and Li, Hanchen and Thakker, Urmish and Zou, James and Olukotun, Kunle},
  journal = {arXiv preprint},
  year    = {2025},
  url     = {https://arxiv.org/abs/2510.04618},
}

@article{yang2026skillopt,
  title   = {{SkillOpt}: Executive Strategy for Self-Evolving Agent Skills},
  author  = {Yang, Yifan and Gong, Ziyang and Huang, Weiquan and Yang, Qihao and Zhou, Ziwei and Huang, Zisu and Li, Yan and Gao, Xuemei and Dai, Qi and Liu, Bei and Qiu, Kai and Yang, Yuqing and Chen, Dongdong and Yang, Xue and Luo, Chong},
  journal = {arXiv preprint},
  year    = {2026},
  url     = {https://arxiv.org/abs/2605.23904},
}

@inproceedings{khattab2023dspy,
  author    = {Khattab, Omar and Singhvi, Arnav and Maheshwari, Paridhi and Zhang, Zhiyuan and Santhanam, Keshav and Vardhamanan, Sri and Haq, Saiful and Sharma, Ashutosh and Joshi, Thomas T. and Moazam, Hanna and Miller, Heather and Zaharia, Matei and Potts, Christopher},
  title     = {{DSPy}: Compiling Declarative Language Model Calls into State-of-the-Art Pipelines},
  booktitle = {Proceedings of the International Conference on Learning Representations (ICLR)},
  year      = {2024},
  url       = {https://openreview.net/forum?id=sY5N0zY5Od},
}

@inproceedings{opsahlong2024mipro,
  title     = {Optimizing Instructions and Demonstrations for Multi-Stage Language Model Programs},
  author    = {Opsahl-Ong, Krista and Ryan, Michael J. and Purtell, Josh and Broman, David and Potts, Christopher and Zaharia, Matei and Khattab, Omar},
  booktitle = {Proceedings of the 2024 Conference on Empirical Methods in Natural Language Processing},
  pages     = {9340--9366},
  publisher = {Association for Computational Linguistics},
  year      = {2024},
  doi       = {10.18653/v1/2024.emnlp-main.525},
}

@article{yuksekgonul2024textgrad,
  title   = {Optimizing Generative {AI} by Backpropagating Language Model Feedback},
  author  = {Yuksekgonul, Mert and Bianchi, Federico and Boen, Joseph and Liu, Sheng and Lu, Pan and Huang, Zhi and Guestrin, Carlos and Zou, James},
  journal = {Nature},
  volume  = {639},
  number  = {8055},
  pages   = {609--616},
  year    = {2025},
  doi     = {10.1038/s41586-025-08661-4},
}

@inproceedings{pryzant2023protegi,
  title     = {Automatic Prompt Optimization with ``Gradient Descent'' and Beam Search},
  author    = {Pryzant, Reid and Iter, Dan and Li, Jerry and Lee, Yin Tat and Zhu, Chenguang and Zeng, Michael},
  booktitle = {Proceedings of the 2023 Conference on Empirical Methods in Natural Language Processing},
  pages     = {7957--7968},
  publisher = {Association for Computational Linguistics},
  year      = {2023},
  doi       = {10.18653/v1/2023.emnlp-main.494},
}

@inproceedings{yang2023opro,
  author    = {Yang, Chengrun and Wang, Xuezhi and Lu, Yifeng and Liu, Hanxiao and Le, Quoc V. and Zhou, Denny and Chen, Xinyun},
  title     = {Large Language Models as Optimizers},
  booktitle = {Proceedings of the International Conference on Learning Representations (ICLR)},
  year      = {2024},
  url       = {https://openreview.net/forum?id=Bb4VGOWELI},
}

@inproceedings{zhou2022ape,
  author    = {Zhou, Yongchao and Muresanu, Andrei Ioan and Han, Ziwen and Paster, Keiran and Pitis, Silviu and Chan, Harris and Ba, Jimmy},
  title     = {Large Language Models Are Human-Level Prompt Engineers},
  booktitle = {Proceedings of the International Conference on Learning Representations (ICLR)},
  year      = {2023},
  url       = {https://openreview.net/forum?id=92gvk82DE-},
}

@inproceedings{zhang2024edc,
  title     = {Extract, Define, Canonicalize: An {LLM}-based Framework for Knowledge Graph Construction},
  author    = {Zhang, Bowen and Soh, Harold},
  booktitle = {Proceedings of the 2024 Conference on Empirical Methods in Natural Language Processing},
  pages     = {9820--9836},
  publisher = {Association for Computational Linguistics},
  year      = {2024},
  doi       = {10.18653/v1/2024.emnlp-main.548},
}

@inproceedings{bai2025autoschemakg,
  title     = {{AutoSchemaKG}: Autonomous Knowledge Graph Construction through Dynamic Schema Induction from Web-Scale Corpora},
  author    = {Bai, Jiaxin and Fan, Wei and Hu, Qi and Zong, Qing and Li, Chunyang and Tsang, Hong Ting and Luo, Hongyu and Yim, Yauwai and Huang, Haoyu and Zhou, Xiao and Qin, Feng and Zheng, Tianshi and Peng, Xi and Yao, Xin and Yang, Huiwen and Wu, Leijie and Ji, Yi and Zhang, Gong and Chen, Renhai and Song, Yangqiu},
  booktitle = {Proceedings of the 64th Annual Meeting of the Association for Computational Linguistics (Volume 1: Long Papers)},
  pages     = {20557--20584},
  publisher = {Association for Computational Linguistics},
  year      = {2026},
  doi       = {10.18653/v1/2026.acl-long.942},
}

@inproceedings{mo2025kggen,
  title     = {{KGGen}: Extracting Knowledge Graphs from Plain Text with Language Models},
  author    = {Mo, Belinda and Yu, Kyssen and Kazdan, Joshua and Mpala, Proud and Yu, Lisa and Cundy, Chris and Kanatsoulis, Charilaos and Koyejo, Sanmi},
  booktitle = {Advances in Neural Information Processing Systems 38},
  year      = {2025},
  doi       = {10.52202/085713-1010},
}

@article{misra2026kgcqual,
  title   = {{KGCQual}: An Interpretable Framework for Evaluating the Knowledge Graph Construction Quality from Text},
  author  = {Misra, Nipun and Udandarao, Vikranth and Gupta, Aanchal and Kumar, Yogender and Mukherjee, Manuj and Mutharaju, Raghava},
  journal = {arXiv preprint},
  year    = {2026},
  url     = {https://arxiv.org/abs/2607.10212},
}

@inproceedings{heist2023kgreat,
  title     = {{KGrEaT}: A Framework to Evaluate Knowledge Graphs via Downstream Tasks},
  author    = {Heist, Nicolas and Hertling, Sven and Paulheim, Heiko},
  booktitle = {Proceedings of the 32nd ACM International Conference on Information and Knowledge Management},
  pages     = {3938--3942},
  publisher = {Association for Computing Machinery},
  year      = {2023},
  doi       = {10.1145/3583780.3615241},
}

@inproceedings{li2024findkg,
  title     = {{FinDKG}: Dynamic Knowledge Graphs with Large Language Models for Detecting Global Trends in Financial Markets},
  author    = {Li, Xiaohui Victor and Sanna Passino, Francesco},
  booktitle = {Proceedings of the 5th ACM International Conference on AI in Finance},
  pages     = {573--581},
  publisher = {Association for Computing Machinery},
  year      = {2024},
  doi       = {10.1145/3677052.3698603},
}

@inproceedings{li2025finkario,
  title     = {{FinKario}: Event-Enhanced Automated Construction of Financial Knowledge Graph},
  author    = {Li, Xiang and Sun, Penglei and Zhou, Wanyun and Wei, Zikai and Zhang, Yongqi and Chu, Xiaowen},
  booktitle = {Proceedings of the 64th Annual Meeting of the Association for Computational Linguistics (Volume 1: Long Papers)},
  pages     = {9826--9845},
  publisher = {Association for Computational Linguistics},
  year      = {2026},
  doi       = {10.18653/v1/2026.acl-long.446},
}

\end{document}